# Evaluating Large Language Models with Tests of Spanish as a Foreign Language: Pass or Fail?

M. Mayor-Rocher, N. Melero, E. Merino-Gómez, M. Grandury, J. Conde and P. Reviriego

**Abstract**

Large Language Models (LLMs) have been profusely evaluated on their ability to answer questions on many topics and their performance on different natural language understanding tasks. Those tests are usually conducted in English, but most LLM users are not native English speakers. Therefore, it is of interest to analyze how LLMs understand other languages at different levels: from paragraphs to morphems. In this paper, we evaluate the performance of state-of-the-art LLMs in TELEIA, a recently released benchmark with similar questions to those of Spanish exams for foreign students, covering topics such as reading comprehension, word formation, meaning and compositional semantics, and grammar. The results show that LLMs perform well at understanding Spanish but are still far from achieving the level of a native speaker in terms of grammatical competence.

1. ## Introduction

The release of ChatGPT[1] at the end of 2022 has spurred a revolution, putting Natural Language Processing (NLP) in the spotlight of the general public (Wu 2023). Today, billions of people interact with Artificial Intelligence tools such as conversational chatbots or agents using natural language. Behind this revolution are Large Language Models (LLMs) that are capable of advanced natural language processing (Santra 2023). These LLMs have billions of parameters and are trained on huge amounts of text to achieve an unprecedented level of performance in most NLP tasks.

There are many LLMs. Some of them such as GPT (Achiam 2024), Claude or Gemini (Anil 2024) are proprietary and can only be used through the user and application programming interfaces provided by the companies that developed the models: OpenAI, Anthropic, and Google, respectively. Instead, in other models such as Llama (Dubey, 2024), Mistral (Jiang 2023) or Yi (Young 2024), the parameters and code are publicly available, which allows users to run them locally. However, the data used for training the models is rarely public. Only in some cases we know the languages present in the dataset and in which percentage they are. The LLM ecosystem is extremely dynamic with new models being released continuously.

As LLMs evolve and new models are released, there is a need to understand their performance on different tasks. The most common method to evaluate LLMs is with multiple choice question tests that cover many aspects, from knowledge on different topics and logical reasoning to solving math problems (Guo 2023). For example, the Massive Multitask Language Understanding (MMLU) benchmark covers 57 different subjects, from math to philosophy or foreign policy (Hendrycks 2021) and newer benchmarks evaluate hundreds of topics and tasks (Srivastava 2024).

---

[1] https://openai.com/chatgpt/

In many cases, the benchmarks take the questions from existing exams, tests, or books; and more recently questions are also specifically written to test LLMs. A limitation of these benchmarks is that they are generally written in English. This leaves out linguistic capabilities relevant to most LLM users, who are not native English speakers, like the more than 600 million potential users in Spain and Latin America (Grandury 2024). To evaluate LLMs in languages other than English, a common practice is to machine translate the questions from the original benchmarks into English, which has limitations (Plaza 2024).

To understand the performance of LLMs in Spanish, specific benchmarks and resources are being developed in the context of the collaborative #Somos600M Project led by SomosNLP (Grandury 2024). Among those, there is a specific test, TELEIA (Mayor-Rocher 2024) designed to evaluate the proficiency of the Spanish language. The test is composed of three independent parts, each of which has questions that are similar to those in three exams: SIELE, Cervantes AVE and PCE, used to evaluate foreign students who are learning Spanish.

In this paper, we present and analyze the results of running TELEIA on a set of eight LLMs. The goal is to build an understanding of the knowledge that LLMs have of Spanish. The results for each of the three parts of TELEIA are manually analyzed to determine whether there are common errors across LLMs and which are the hardest questions for them. The results obtained provide valuable information on the limitations of state-of-the-art LLMs in Spanish and also point to possible explanations of those errors that could potentially be used to improve the performance of future LLMs in Spanish.

The rest of the paper is organized as follows. Section 2 describes the methodology used for evaluation in terms of the LLMs selected, the benchmark used and the actual procedure to run the test. The detailed results are presented in section 3 and discussed in section 4. The paper ends with the conclusion in section 5.

## 2. Methodology

This section briefly describes the benchmark and the LLMs used in the evaluation as well as the procedure and tools used to run the test.

**2.1 Benchmark**

As mentioned in the introduction, the TELEIA dataset (Mayor-Rocher 2024) is composed of questions that are similar to those in three Spanish exams for non-native speakers:

- PCE: an exam that allows access to university for those students who have not followed the general Spanish university access system (UNED 2023).

- SIELE (Servicio Internacional de Evaluación de la Lengua Española): a unique multi-level exam to certify the level of proficiency in the Spanish language (Soria 2016).

- Cervantes AVE ("Spanish Virtual Classroom"): an official online placement test designed for the self-assessment of the use of Spanish by non-native speakers (Juan-Lázaro 2009).

In total, the test has 100 questions, 34, 36 and 30 for each of the categories. The first part, SIELE, focuses on reading comprehension, while Cervantes AVE and PCE focus on grammatical competence. Additional details for each of the parts will be provided in the next section when discussing the results.

## 2.2 LLMs

The goal of the evaluation is to understand the LLMs' proficiency in Spanish, so the set of LLMs evaluated should be representative of the different LLMs and at the same time not too large, so that the manual analysis of the errors can be made with a reasonable effort. To achieve a trade-off between the two conflicting requirements, eight conversational LLMs have been evaluated. On the proprietary side, two versions of ChatGPT, 3.5 and 4, have been selected. For open LLMs, our selection includes Llama, Mistral and Gemma models that are focused on English, Yi which includes Chinese, and an Occiglot model that is designed to support the main European languages covering Spanish as one of their target languages. This provides a set of LLMs developed by different companies, with different sizes and optimized for different languages, so that the results can be representative of the LLM landscape. Since the task we are going to evaluate the models on is multiple choice question answering, we have selected the instruction-tuned or chat-like versions of the models, i.e. the result of adapting the pre-trained LLM to follow instructions by fine-tuning it with input-output (or question-answer) pairs (Longpre 2023).

## 2.3 Evaluation procedure

For the open models, the questions in the TELEIA test were run locally on NVIDIA GPUs using the LLM harness (Sutawika 2024). For the proprietary models, a script that calls OpenAI's Application Programming Interface (API) was used to automate the process ensuring that the same prompts are used. The answers were logged in files that are publicly available at Zenodo[2].

## 3. Results and Analysis

In this section, the results for each of the parts of the benchmark are presented and analyzed.

### 3.1 PCE

The PCE exams of UNED (National University of Distance Education) allow access to university for those students who have not followed the general Spanish university access system. One of the tasks in these exams is a Spanish test with questions related to a text. From the different models available online, questions on morphology and semantics have been selected to create 34 new questions for TELEIA. Thus, the questions inspired by the

---

[2] Zenodo TELEIA results: https://zenodo.org/doi/10.5281/zenodo.13643393

PCE exams are divided into two balanced groups: those related to word formation (17/34), as in example (1), and those pertaining to the field of semantics, both meaning and compositional (17/34), as in example (2):

(1) Las palabras "vertebral", "estomacal" y "dorsal" son:
   a) Derivados a partir de base sustantiva (correct answer).
   b) Extranjerismos adaptados gráficamente al español.
   c) Derivados a partir de una base adjetiva.

   The words "vertebral", "stomachal," and "dorsal" are:
   a) Derived from a noun base (correct answer).
   b) Foreign words graphically adapted to Spanish.
   c) Derived from an adjective base.

(2) Las palabras "médico", "profesor" y "carpintero" mantienen entre ellas una relación de:
   a) Hiperonimia.
   b) Sinonimia.
   c) Cohiponimia (correct answer).

   The words "doctor", "teacher" and "carpenter" share a relationship of:
   a) Hypernymy.
   b) Synonymy.
   c) Co-hyponymy (correct answer).

   El sintagma "un diluvio de solidaridad" es:
   a) Una metonimia.
   b) Un coloquialismo.
   c) Una metáfora (correct answer).

   The phrase "un diluvio de solidaridad" (lit. a flood of solidarity) is:
   a) A metonymy.
   b) A colloquialism.
   c) A metaphor (correct answer).

The accuracy rates of the models for the questions on the original PCE exams and TELEIA test are as follows:

| Model | Original's accuracy rate | TELEIA's Accuracy rate |
|---|---|---|
| GPT 4 | 82.35% | 79.41% |
| Meta-Llama-3-8B-Instruct | 70.59% | 70.59% |
| GPT 3.5 | 61.76% | 64.71% |
| Yi-6B-Chat | 61.76% | 64.71% |
| Gemma-7b-it | 47.06% | 55.08% |
| Mistral-7B-Instruct-v0.1 | 52.94% | 50.00% |
| Occiglot-7b-es-en-instruct | 52.94% | 44.12% |
| Llama-2-7b-chat-hf | 32.36% | 32.35% |

TABLE 1. Accuracy rates of different models for PCE

As shown in Table 1, most of the models (6/8) pass this exam. In TELEIA, GPT-4 achieves the highest accuracy rate, with 79.41%. This is the best model, capable of identifying word formation processes, especially inflectional ones, and understanding the semantic meaning provided by these morphemes. Additionally, it very accurately identifies semantic relationships such as hypernymy, hyponymy, antonymy, and synonymy, and understands the compositional meaning of certain phrases and metaphors. It also detects acronyms like TIC (Tecnologías de la Información y las Comunicaciones / Information and Communication Technologies) and, unlike in the original test, recognizes foreign terms like the English terms *email*, *chat*, and *bot*. The same applies to the next best model, Meta-Llama-3-8B-Instruct, which is effective in detecting foreign terms and acronyms. However, this model makes more errors in questions related to word formation and, on the semantic level, with compositional processes such as the metaphor of *burbuja* in "burbuja inmobiliaria" (housing bubble).

GPT-3.5 and Yi-6B-Chat both achieve the same accuracy rate of 64.71% in TELEIA (and also the same rate of 61.76% in the original test). However, the former is slightly worse than the latter in detecting word formation processes, as the latter can detect patterns among series of words and derivation and parasynthesis processes that the former fails to identify. In any case, both are satisfactory in the semantic domain, as they can establish synonymy and antonymy relationships, although GPT-3.5 recognizes meanings of words and compound expressions that Yi-6B-Chat sometimes does not. Both models have difficulties establishing relationships between words of the same family, such as hypernymy and co-hyponymy.

On the other hand, Gemma-7b-it and Mistral-7B-Instruct-v0.1 fluctuate around the passing mark in TELEIA, although Gemma-7b-it failed the original test with 47.96%. Both models struggle with identifying morphological processes of prefixation, suffixation, and composition, as well as semantic relationships between terms. Semantically, Mistral-7B-Instruct-v0.1 does not adequately identify synonyms, metaphors, or metonymies, such as *clínex* (*Kleenex*, tissue). Both models seem capable of identifying foreign terms and acronyms. Finally, both Occiglot-7b-es-en-instruct and Llama-2-7b-chat-hf fail the test (Occiglot had passed the original test). These models have trouble identifying word formation patterns as well as meanings and relationships between words of the same family. On this occasion, Llama-2-7b-chat-hf did not recognize foreign terms.

It has been observed, from highest to lowest, the ability of the models to establish morphological and semantic relationships, with the latter being better identified by the models. For example, this is evident as all models correctly identify the meaning of the phrase "ser alguien de pocas palabras" ("to be a man of few words"). It is notable that all models correctly answer the question about the formation of the word "TIC" (Information and Communication Technologies) through acronymy, which is common in the tech world. However, all models fail—and also failed in the original test—a question about the neologisms "cafedemia" (*café* + *pandemia* ~ coffee + pandemic) and "cafexicación" (*café* + *intoxicación* ~ coffee + intoxication), clearly indicating that models are currently incapable of processing this type of data that humans can naturally identify.

### 3.2 SIELE

The Spanish test of SIELE (Servicio Internacional de Evaluación de la Lengua Española) consists of a unique multi-level exam to certify proficiency in the Spanish language. The exam is divided into four sections: reading comprehension, listening comprehension, written expression and interaction, and oral expression and interaction. To test the functioning of different models, a similar test to the reading comprehension section published by SIELE[3] has been developed. For this purpose, language samples were drafted to be tested in different models: four texts, two long ones ranging from 300 to 470 words, and two short ones of about 60-80 words. As in the SIELE exam, these texts belong to various common discursive genres: interviews, emails, and instructions for attending a theatrical performance.

To verify that the characteristics of the tests drafted for this analysis (included in the TELEIA database) and the original SIELE tests have analogous features, the aggregated data of both are presented in Table 1. In general, the models respond with a similar percentage of correct answers in the case of the written comprehension test (SIELE) and the ones drafted for this study (TELEIA). It is interesting to note that in the models where higher accuracy rates are observed—Meta-Llama-3-8B-Instruct, ChatGPT3.5, and ChatGPT4—the results are very close or identical.

| Model | Original's SIELE accuracy rate | TELEIA's (exams based on SIELE's formats) accuracy rate |
|---|---|---|
| ChatGPT4 | 97.22% | 97.22% |
| ChatGPT3.5 | 86.11% | 91.67% |
| Meta-Llama-3-8B-Instruct | 88.89% | 88.89% |
| Gemma-7b-it[4] | 55.56% | 77.78% |
| Mistral-7B-Instruct-v0.1 | 66.67% | 77.78% |
| Occiglot-7b-es-en-instruct | 66.67% | 77.78% |
| Yi-6B-Chat | 55.56% | 66.67% |
| Llama-2-7b-chat-hf[5] | 58.33% | 55.56% |

TABLE 2. Comparison of the accuracy rates of different models for the original SIELE exams and those specifically developed for this study (TELEIA).

There are three questions where six out of the eight models made mistakes. In the first text, three people are interviewed and asked to recall a trip they took as children. The three interviews total 469 words, and 12 questions are based on them. Out of the 96 possible valid responses (8 models x 12 questions), 28 errors were identified. This 29% error rate is distributed without an apparent pattern, except for the question: "¿Quién ha cambiado de opinión acerca de la catedral tras volver de mayor?" ("Who changed their opinion about the cathedral after returning as an adult?") In this case, only the ChatGPT3.5 and ChatGPT4 models provided the correct answer. Of the six incorrect answers, five mistakenly pointed to option 'c', likely because a cathedral was also mentioned in that part of the text. The correct answer was implicitly contained in the fragment: "Pensé que la catedral era el edificio más inmenso que había visto en mi vida. Ya de adulto, he regresado y no me ha parecido tan grande" ("I thought the cathedral was the most immense building I had ever seen. As an adult, I returned and it didn't seem as large"). The fact that the subject was implicit—"[the cathedral or the building] didn't seem as large"—which had to be inferred from the previous sentence, may have caused a coreferential error (Jurafsky & Martin 2023), given the separation first by a paragraph break and second by the distant positioning of the second part, which refers to the opinion "it didn't seem [as large]." The error could also be due to the failure to identify the semantic correlation between the sequence "cambiado de opinión" ("changed their opinion") and "no me ha parecido [ya de adulto] tan grande" ("didn't seem [as large] [as an adult]").

---

[3] https://siele.org/documents/10180/84852/SIELE+Modelo+0/c2073785-5535-443b-969a-500e425ac06c

[4] In additional testing conducted with the successive version of Gemma-7b-it: Gemma-9B, the result was a 94.44% accuracy rate, improving by more than 16 percentage points compared to the previous version.

[5] In the test conducted with the successive version of Llama: Llama 3.1, the result for the TELEIA test achieved a 100% accuracy rate, improving by more than 44 percentage points compared to the previous version.

The text with the highest percentage of errors is the second one, which consists of a brief text:

> "Hola, Andrea:
>
> Soy Marina, la compañera de trabajo de Juan. No sé si sabes que han destinado a su mujer a Santander y que el viernes que viene será su último día de trabajo. Hemos pensado recoger dinero para hacerle un regalo entre todos. Necesito que me digas si quieres participar en el donativo para el regalo y si podrás estar en el pequeño homenaje que le haremos el próximo jueves en la oficina.
>
> Espero tu respuesta.
>
> Saludos.
>
> Marina"

The text, consisting of only 79 words, is evaluated through 6 questions. If all eight models had provided valid answers, there would have been 48 correct responses. However, 16 errors were recorded, resulting in a 33% error rate in a text whose complexity lies in the relational aspects between the different participants. The question that accumulated the most errors is as follows:

> En el texto, Marina…
> a) dice que han pensado en hacer un regalo a Andrea entre todos los compañeros.
> b) dice que van a poner dinero para hacerle un regalo a la mujer de Juan.
> c) dice que entre todos le van a hacer un regalo a Juan.

Only Occiglot-7b-es-en-instruct and ChatGPT4 provided the correct answer, 'c'. Of the six incorrect answers, five indicated that the correct answer was 'a', meaning they assumed that *Andrea*, the recipient of the message, was also the intended recipient of the gift. There could be perceived ambiguity in the phrase "Hemos pensado recoger dinero para hacerle un regalo entre todos" if it were interpreted that Marina, the sender, was addressing Andrea formally (that is, assuming that the enclitic pronoun "-le" is using the third person, resulting in an ambiguity). However, the context, right from the beginning, consistently shows the use of informal language ("tuteo"), which should resolve any such ambiguity. This frequency of errors in such a short text is undoubtedly also attributable to referential difficulties, similar to the massively incorrect response to the question in the first text.

Regarding the third text, which consists of 316 words, 12 questions were asked. Out of the 96 correct answers that could have been recorded, 18 contained errors, resulting in an error rate of 18.8%. The highest number of errors (12) occurred in the following two questions:

En el equipo de Julia…
a) hay dos arqueólogos.
b) hay un arqueólogo y un marinero.
c) no hay gente para trabajar en verano.

The correct answer is 'a'. This solution can be deduced from the periphrasis in the explanatory phrase that accompanies "Nicola and Gianni" in the text: "Nosotros trabajamos con Nicola y Gianni, que son expertos en arqueología". However, the six models that made errors did so uniformly, indicating that the correct answer was 'b'. Only Yi-6B-Chat and, once again, ChatGPT4 provided the correct answer. The incorrect answer chosen by the models, "hay un arqueólogo y un marinero" does not seem to stem from any ambiguity in the text. The word *marinero* does not appear, and the need for "una persona que nos pueda ayudar con las actividades relacionadas con la navegación" which could be assimilated with the meaning of *marinero* is found after the adversative connector *pero*.

The other question that resulted in 6 errors, also from the third text, is as follows:
Julia avisa a Carmen de que…
a) los mosquitos en la zona de trabajo pican solo por la noche.
b) hay muchos mosquitos.
c) hay pocos mosquitos, pero pican a cualquier hora del día.

The correct answer is 'b'. Only ChatGPT 3.5 and ChatGPT4 got it right. The rest of the models failed all pointing to answer 'c'. The segment of the text that refers to the mosquitoes, which is the focus of the question, is as follows: "Para mí lo peor son las picaduras de los mosquitos. Están por todas partes y pican a cualquier hora del día. No es como en Sevilla, que solo pican por la noche". The meaning of "there are many mosquitoes". corresponding to option 'b', is conveyed indirectly but unequivocally through the expression "están [los mosquitos] por todas partes". However, the six models that made mistakes chose the incorrect answer, "there are few mosquitoes", likely because they detected the identical six-word sequence "pican a cualquier hora del día" both in the question and in the body of the text. ChatGPT, in both versions, correctly identified that the sequence "bite at any time of the day" is preceded by an adversative conjunction and by a clause that invalidates the interpretation leading to the wrong answer.

| TELEIA Texts (similar to the SIELE reading comprehension test) | Number of words | Number of questions | Number of errors | Error percentage considering all models |
|---|---|---|---|---|
| Text 1. Interviews about a childhood trip: | 469 | 12 | 28 | 29% |
| Text 2. Message to buy a gift for a coworker: | 79 | 6 | 16 | 33.3% |
| Text 3. Message to request a friend's collaboration on a work project: | 316 | 12 | 18 | 18.8% |
| Text 4. Theatre play advertisement for the Mérida Festival: | 64 | 6 | 1 | 0.5% |

TABLE 3. Error percentage for the four texts

The fourth text, consisting of only 64 words, has the highest accuracy rate. Six questions were asked, which, when multiplied by the 8 models tested, would result in a total of 48 responses. In this segment, only one error was made, resulting in an error rate of just 0.5%. The text systematically presented information on how to attend a theatrical performance:

> "Lugar: Teatro Romano de Mérida
> Obra: Edipo Rey
> Compañía: Ballabina Teatre
> Fecha: sábado, 3 de agosto de 2024
> Hora: se puede entrar desde las 19:30. La representación empieza a las 20:00 horas.
> Duración: 80 minutos
> Precio: desde 5 €.
> Edad: mayores de 12 años
> Adquiere tu entrada en: https://www.entradas.com/artist/festival-de-merida/edipo-rey-1963930/?affiliate=ADE
> Disponibilidad de asientos adaptados para personas con movilidad reducida
>
> Información adicional: 924 123 456 789"
>
> The only error identified is found in the response to the first question:
>
> "En la información se dice que…
> a)   la obra de teatro empieza a las 20:00
> b)   es una obra para niños
> c)   las entradas se compran por internet",

where the correct answer is 'c', but the Occiglot model responded with 'b'. In this case, it seems that the models easily pick up on the systematized data that is not linked by any discourse connectors. Referential aspects, which were likely the cause of the high error rates in the previous texts, are not present in this model.

In conclusion, in the first text, some relational difficulties arise from dealing with the testimonies of three people, although deliberate semantic and lexical similarities were introduced to assess their comprehension. In the second text, despite its brevity, it identifies not only the sender and recipient of the gift but also the recipient's wife and the recipient's work colleagues. With such a limited number of words, complex relationships among many individuals are presented, contributing to the highest percentage of errors. In the third text, despite being nearly four times longer than the second text, the number of errors is lower. At least one of the two major errors can be explained by the identification of a significant 6-word sequence in both the question and the text. The simplicity of character relationships, identified by proper names (Carmen, Julia, Nicola, Gianni), results in fewer difficulties in identifying the correct answers. Finally, the near absence of errors in the last text, which is almost the same length as the second text that accumulated the most errors, is related to its syntactically simple presentation of information, making it easily retrievable and free of apparent ambiguity.

## 3.3 AVE

The AVE test[6] ("Spanish Virtual Classroom") is an official online placement test. It was developed in 2011 by the Instituto Cervantes, the institute created by the Spanish government for the teaching and promotion of the Spanish language around the world. The AVE test is public and was designed for the purpose of self-assessment of the use of Spanish by non-native speakers. It consists of three parts: a grammar test with 30 items, a reading comprehension exercise and a listening comprehension exercise. The grammar section presents a traditional, fill-in-the-gap multiple-choice test assessing the use of adjectives (2 items), adverbs (1), conjunctions (1), prepositions (3), pronouns (7) and verbs (16).

TELEIA includes a set of 30 items based on that section, and tests similar skills. The choices presented to the assessee were designed considering the most frequent grammatical mistakes made by learners of Spanish as a foreign language: the difference between the past tenses, the use of prepositions and pronouns, the combination of complex verbal modes (subjunctive versus indicative), etc.

Whereas the PCE test assesses the morphosyntactic and semantic knowledge of Spanish from a linguist's perspective, SIELE and AVE are language tests that any native Spanish speaker with an average cultural background could pass with a 100% score. However, this doesn't seem to be the case with non-human users of the language. Below is a table with the accuracy rates of the models for the original AVE test and the corresponding section in TELEIA.

| Model | AVE's accuracy rate | TELEIA's Accuracy rate |
| --- | --- | --- |
| *GPT 4* | 80% | 80% |
| *GPT 3.5* | 50% | 73.3% |
| *Meta-Llama-3-8B-Instruct* | 46.6% | 50% |
| *Yi-6B-Chat* | 43.3% | 40% |
| *Mistral-7B-Instruct-v0.1* | 33.3% | 33% |
| *Occiglot-7b-es-en-instruct* | 43.3% | 43.3% |
| *Gemma-7b-it* | 46.6% | 36.6% |
| *Llama-2-7b-chat-hf* | 23.3% | 23.3% |

TABLE 4. Accuracy rates of the different models for AVE and TELEIA.

The performance of most of the LLMs tested (6 out of 8) does not differ significantly in both tests but does vary across models. Chat GPT4 and GPT 3.5 reach higher accuracy rates than the rest of the models in both, AVE and TELEIA; whereas Mistral-7B-Instruct-v0.1 and Llama-2-7b-chat-hf reach the lowest score in both tests. Only GPT 4 and GPT 3.5 pass both tests (scoring >50%), and Meta-Llama-3-8B-Instruct reaches 50% only in the TELEIA test. The rest of the models perform rather poorly (scoring <50%), reaching similar accuracy rates in both tests.

As shown in the table, results are very similar in both tests for any of the models, except for GPT 3.5 and Meta-Llama-3-8B-Instruct, which perform better in the TELEIA test (73.3% versus 50%, and 50% versus 46.6% respectively). Only Yi-6B-Chat and Gemma-

---

[6] The AVE test can be accessed here: https://pruebadenivel.cervantes.es/exam.php?id=17

7b-it score lower in the TELEIA test than in the original AVE test (46.6% versus 50%, and 36.6% versus 46.6% respectively).

The items of the tests in which most LLMs provide incorrect answers do not follow a pattern and do not coincide in both tests. However, the types of questions that posed the highest difficulty for the models tested were as follows in both tests (with only 3 or fewer models out of 8 answering accurately):

- Type of question: adjectival. Mandatory elision of the final vowel in some ordinal numbers and adjectives when these precede the noun ("primer"/"primero", "mal"/"malo").
- Type of question: prepositional. Use of prepositions in specific word combinations: "Estar de pie", "viajamos por todo el país".
- Type of question: verbal. Combination and difference of the preterites (Spanish present perfect versus simple past: "has estado alguna vez"/ "la semana pasada vi…"; imperfect versus simple past: "ayer vi que había", "estaba sola y oí"; past perfect versus simple past: "ya había terminado cuando llamaron").
- Type of question: verbal. Difference between the two forms of *to be* in Spanish, *ser* and *estar*, when referring to a happening or event: "Dónde es la fiesta", "la conferencia es en…".
- Type of question: verbal. Use of the subjunctive mode: "me da rabia que me digas…" / "me pone nerviosa que la gente hable en el cine", "creo que hará mal tiempo, pero no creo que llueva", "pienso que estás equivocado: no creo que las cosas sean como tú piensas", "buscan a alguien que sepa…".
- Type of question: verbal. Use of verbal periphrasis: "llevar [tiempo] sin hacer", "sigues yendo", "sigues aprendiendo".

This implies that, according to the answers given by most of the models tested, ungrammatical sentences as the following (3a-h) should be correct (among many others):

(3) a) *Ha ganado el primero premio.
 b) *No creo que las cosas son como tú piensas.
 c) *Dónde hay la fiesta?
 d) *La conferencia está en el salón de actos.
 e) *Hará mal tiempo, pero no creo que lloverá.
 f) *Sigues a aprender a tocar la guitarra.
 g) *Fui meses sin ir a clase.

All of the above mistakes are also very frequent in human learners of Spanish as a foreign language (Madrid 1999). These could be attributed to a negative language transfer; i.e., an interference from the learner's mother tongue (especially if this is English, since many of these ungrammatical segments could be literal translations from the English with incorrect verb choices). This is consistent with observations from recent works on the multilingual capabilities of LLMs (Terryn 2024; Wendler2024) and the fact that English dominates the training datasets of most LLMs accounting in most cases for more than 90% of the text used for training.

It is worth noting, however, that the LLMs did not produce these incorrect sentences themselves as full language units. Instead, they chose the wrong answers from a set of isolated words. Thus, the meaning of the chosen words is correct, but their form (conjugation, verb choice, etc.) becomes incorrect when inserted into the language chain, since certain combinations require specific choices that are mandatory from the grammatical point of view. These combinations (and their exceptions) are usually at the centre of Spanish classes for human learners, and therefore appear frequently in this type of grammar tests.

We may conclude that the most frequent mistakes across models are very similar to those that a human learner of Spanish could make, with none of the models reaching a C1 or C2[7] level of proficiency in the language (at least when completing this type of grammar test).

### 3. 4 Open LLM Leaderboard

We know that the majority of the data used to train these models is in English. In the corresponding model papers, only Meta-Llama-3-8B-Instruct and Occiglot-7b-es-en-instruct explicitly mention that they include Spanish text in their training data. This could explain why these are the two best-performing open models at SIELE and AVE, although this doesn't apply to the PCE test.

In the previous sections, we have seen that LLMs' performance in reading comprehension (SIELE) is significantly higher than in morphology, semantics (PCE) and grammar (AVE). This could be explained by the fact that grammar depends much more on the language than some natural language understanding abilities like information extraction. Even if their training data didn't include a high percentage of Spanish, the instructions used during the supervised fine-tuning step of the models probably included abstractive question-answering tasks, i.e., questions to answer given a context. The knowledge acquired might be transferable to other languages, unlike specifics about Spanish grammar.

To analyze this correlation we are going to compare the performance of the open LLMs on TELEIA with their performance on the Open LLM Leaderboard (Fourrier 2024), the most common automated evaluation (vs. human evaluation) benchmark for English tasks. It evaluates general and scientific knowledge, logic and math reasoning, instruction following, and common sense text understanding. We see in Table 5 that, on average, the models follow the same order, which might hint to the fact that understanding well English helps in understanding Spanish.

| Model | TELEIA's Accuracy rate | Open LLM Leaderboard |
|---|---|---|
| *GPT 4* | 80% | NA |
| *GPT 3.5* | 73.3% | NA |
| *Meta-Llama-3-8B-Instruct* | 50% | 23.91% |
| *Yi-6B-Chat* | 40% | 14% |
| *Mistral-7B-Instruct-v0.1* | 33% | 12.83% |
| *Occiglot-7b-es-en-instruct* | 43.3% | NA |
| *Gemma-7b-it* | 36.6% | 12.67% |
| *Llama-2-7b-chat-hf* | 23.3% | 9.4% |

TABLE 5. Accuracy rates of the different models for TELEIA and for the average of the Open LLM Leaderboard tasks (note that results for the Leaderboard are not available for some models).

---

[7] C1 and C2 are the most advanced levels in the use of a foreign language according to the Common European Framework of Reference for Languages.

## 4. Discussion

After presenting and analyzing the results for each of the parts of TELEIA, it is worth discussing the overall results and summarising the main trends observed. This is done at two levels, first we look at the differences among models and tests to then go deeper and analyze the results for each of the tests.

The main observations on the models from the results at an aggregated level are:

1) No model achieves 100% of correct answers in any of the tests.
2) ChatGPT4 is the best-performing model and open models have significantly worse results with five out of six failing at least one of the tests.
3) The open model average results follow the same order as their corresponding average results in the Open LLM Leaderboard.
4) There is a significant performance increase in Llama models, with the third version nearly duplicating the average results achieved by the second one.

These observations show that current LLMs have limitations when using Spanish and do not achieve the level of native speakers, who are expected to answer correctly all the questions in SIELE and Cervantes AVE. However, the results also show that performance is improving.

The main observations on the tests from the results at an aggregated level are:

1) Performance is significantly better for SIELE which focuses on reading comprehension.
2) Performance is lower for PCE and especially for Cervantes AVE which focuses on different aspects of grammatical competence.

This shows that LLMs are better at understanding text than at grammar. This might be because grammar is more heavily dependent on the specifics of a language, while some natural language understanding tasks, like information extraction, are less reliant on these linguistic nuances.

Turning to the results for each of the tests, the main observations are:

1) Reading comprehension performance is better for structured text that uses simpler grammatical structures.
2) Errors in reading comprehension are in many cases related to references within the text that are implicit or distant.
3) Most models can establish morphological and semantic relationships between words relatively well.
4) All models are unable to identify neologisms.
5) The models have limitations in identifying the specifics of verbal forms in Spanish (subjunctive, verbal periphrasis, forms of *to be* [*ser* o *estar*], preterites) as well as in using prepositions and adjectives.
6) In many cases, the errors made by the models are similar to those commonly made by native English speakers who are learning Spanish (Madrid 1999).

These observations provide specific points for improvement of LLMs and also suggest that LLMs to some extent behave like English-speaking students who are learning

Spanish. This may be due to the dominance of English in the training datasets of most LLMs and is consistent with other works (Wendler 2024).

As an overall summary, LLMs show good performance at understanding Spanish but are still far from achieving the level of a native speaker. Interestingly, the fact that LLMs tend to make the same mistakes as English-speaking students who are learning Spanish, means that the use of questions similar to those in exams for foreign students is pertinent. This further validates the design of the TELEIA benchmark.

## 5. Conclusion

The different tests that make up TELEIA, taken from the PCE, SIELE, and Cervantes - AVE exams, show high accuracy rates when run through the various models, similar to the accuracy rates of the original tests, which proves the adequacy of our adaptation.

The evaluated tests include a series of questions about the Spanish language: morphology, semantics, and reading comprehension. It has been found that the models with the highest number of correct answers are Chat GPT 4, Chat GPT 3.5, and Meta-Llama-3-8B-Instruct, while those with the fewest correct answers include Llama-2-7b-chat-hf and Occiglot-7b-es-en-instruct.

It is observed that the models establish morphological and semantic relationships between words relatively well and can answer reading comprehension questions based on various texts provided to the models. However, the models demonstrate less skill in grammar-related questions in Spanish and are unable to identify neologisms. Additionally, it is noted that they sometimes have difficulty identifying and correlating references within the text.

Finally, a linguistic bias is detected, similar to that found in learners of Spanish as a foreign language. That is, the primary training language of the models, especially English, may interfere with choosing the correct answer within its linguistic context.

## Acknowledgements

This work was supported by the FUN4DATE (PID2022-136684OB-C21/C22) and ENTRUDIT (TED2021-130118B-I00), projects funded by the Spanish Agencia Estatal de Investigacion (AEI) 10.13039/501100011033, by the European Union Chips Act Joint Undertaking project SMARTY (Grant no. 101140087) and by the OpenAI Researcher Access Program. The evaluation was also done in part with equipment that was donated by NVIDIA to support our research.